\documentclass[10pt]{article}

\usepackage[a4paper,left=1.9cm,right=1.9cm,top=2.0cm,bottom=2.2cm]{geometry}

\usepackage{iftex}
\ifPDFTeX
  \usepackage[T1]{fontenc}
  \usepackage[utf8]{inputenc}
  \usepackage{newtxtext,newtxmath}
  \usepackage[scaled=0.92]{helvet}   
\else
  \usepackage{fontspec}
  \usepackage{newtxtext,newtxmath}
\fi
\newcommand{\sans}{\sffamily}

\usepackage{graphicx}
\usepackage{xcolor}
\usepackage[most]{tcolorbox}
\usepackage{tikz}
\usetikzlibrary{positioning,arrows.meta,calc,fit,backgrounds}
\usepackage{booktabs}
\usepackage{array}
\usepackage{ragged2e}
\usepackage{enumitem}
\usepackage{microtype}
\usepackage{colortbl}
\usepackage{wrapfig}  

\definecolor{scnavy}{HTML}{155050}
\definecolor{scblue}{HTML}{004747}
\definecolor{sclink}{HTML}{00A3A3}
\definecolor{scbox}{HTML}{EEF1F5}     
\definecolor{scrule}{HTML}{CDD9E5}
\definecolor{scgray}{HTML}{8A97A6}
\definecolor{scrowalt}{HTML}{F4F7FA}

\usepackage[colorlinks=true,linkcolor=scnavy,urlcolor=sclink,citecolor=sclink]{hyperref}

\usepackage{titlesec}
\titleformat{\section}{\sans\large\bfseries\color{scnavy}}{\thesection}{0.6em}{}
\titleformat{\subsection}{\sans\normalsize\bfseries\color{scblue}}{\thesubsection}{0.6em}{}
\titlespacing*{\section}{0pt}{14pt}{5pt}
\titlespacing*{\subsection}{0pt}{9pt}{3pt}

\newcolumntype{L}[1]{>{\raggedright\arraybackslash}p{#1}}
\newcolumntype{R}[1]{>{\raggedleft\arraybackslash}p{#1}}

\newcommand{\thd}[1]{\textbf{\textcolor{white}{#1}}}      
\newcommand{\hdrow}{\rowcolor{scnavy}}

\usepackage{fancyhdr}
\begin{document}


\begin{tcolorbox}[
    enhanced, colback=scbox, colframe=scbox, boxrule=0pt,
    arc=7pt, left=16pt, right=16pt, top=15pt, bottom=13pt,
    drop shadow={black!12}]

{\sans\bfseries\fontsize{20}{24}\selectfont \color{scnavy}
COLIP-2: Olfaction-Vision-Language Embeddings \par}
\vspace{5pt}
{\sans\large\color{scblue} External Model Card\par}
\vspace{6pt}


Kordel K. France\\
\textit{Scentience, Inc.}
\newcommand{\orcidauthorA}{0000-0001-6682-5337}
\vspace{8pt}

\hrule height 0.4pt \vspace{8pt}
\textbf{Abstract.}\;
The \textit{Contrastive Olfaction-Language-Image Pre-training 2 (COLIP-2)} model is a multimodal embeddings space that places olfaction as a first-class citizen among vision and language.
Molecular structure, gas-sensor readings, odor-descriptor language, and images are all trained into a single shared representation space, so that a robot can localize a detected aroma to objects in a scene probabilistically.
No \textit{ImageNet}-scale datasets of paired image-scent examples exists which warrants the need for their collection.
Our intent with the release of \textit{COLIP-2} is to demonstrate the limit of what can be built for robotics with open-sourced olfactory data in order to ground the argument for why new methodologies and datasets are necessary in order to enable advanced olfactory-oriented perception capabilities.
We enumerate results from internal testing of the \textit{COLIP-2} architecture and make necessary optimizations to run the model at the edge for real-time robotics applications. 
While developed with robotics in mind, the design of \textit{COLIP-2} has been influenced by experts across many disciplines of science in academia and industry, and we hope that the model can be useful in any multimodal domain requiring olfactory intelligence.

\vspace{9pt}

\hrule height 0.4pt \vspace{7pt}
{\setlength{\parskip}{2pt}
\textbf{\sans Date:}\; 18 July 2026 \hfill\;\\
\textbf{\sans Version:}\; v2.0.1 (external release) \\
\textbf{\sans Model type:}\; Multi-modal (olfaction-vision-language) embeddings \\
\textbf{\sans Correspondence:}\; \href{mailto:kordel@scentience.ai}{info@scentience.ai} \\
\textbf{\sans Access:}\; \href{https://scentience.ai/ai-models}{Scentience API} \\
\textbf{\sans Intended use:}\; Robotics research (see \S\ref{sec:license}) \\
}

\vspace{-50pt}
\hfill\raisebox{0pt}{\includegraphics[height=40pt]{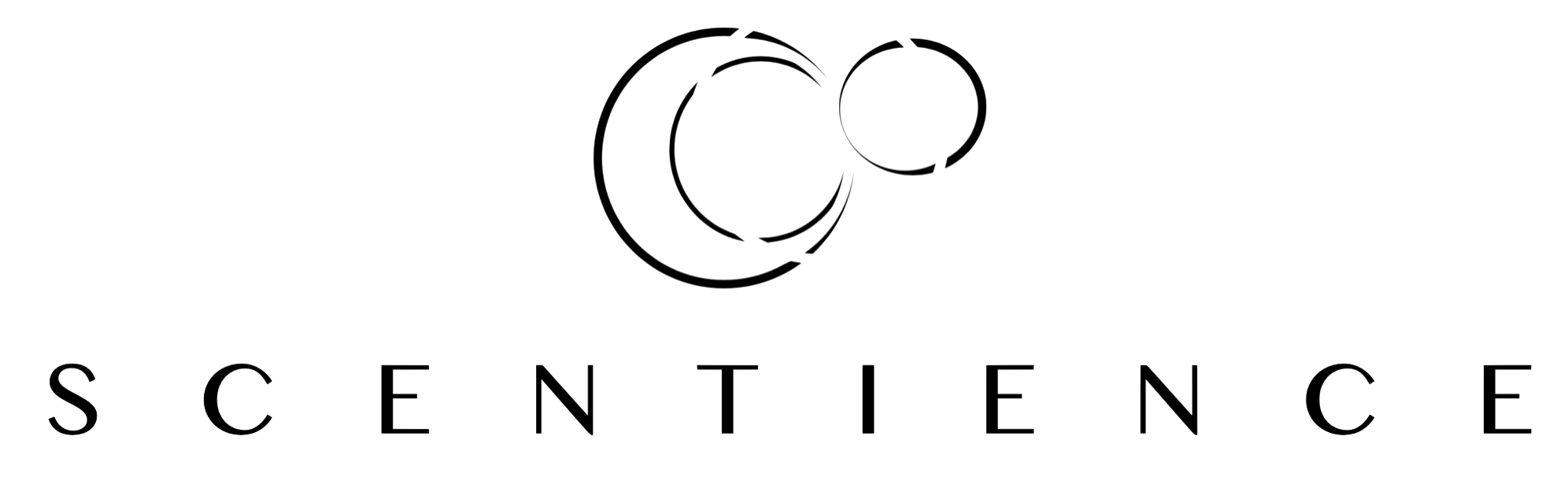}}
\vspace{5pt}

\end{tcolorbox}

\vspace{6pt}

\section{Overview}

Existing foundation models lack a robust sense of olfaction and molecular interpretability.
Scentience's \textit{Contrastive Olfaction-Language-Image Pre-training 2 (COLIP-2)} model connects three perceptions that are rarely modeled together inside one embedding space: a molecule's structure, visual representations that emit said molecule, and the aroma descriptors with which the molecules associates. 
Similar to the foundational \textit{CLIP} \cite{radford2021clip} and \textit{SigLIP} \cite{zhai2023siglip}, \textit{COLIP-2} uses contrastive loss to align three modalities together.
\textit{COLIP-2} is an improved version of the previously released \textit{COLIP-1} model \cite{france2025diffusiongraphneuralnetworks, france2025colip-ovlembeddings}, containing architectural improvements, a cleaner training regime, and additional encoders that enable the model to be easier integrated into robotics platforms.
While the detailed architecture for \textit{COLIP-2} is closed to Scentience, we release this model card in order to summarize the model's design intent, capabilities, evaluation, and limitations for external audiences in order to encourage more research into olfactory robotics. 

The objective of \textit{COLIP-2} is to enable any robot or device with a camera and an olfaction sensor to point at a scene, detect an
aroma, and answer the question \emph{``from which objects in this scene are the observed molecules or aromas probably coming?''}.
Enabling this singular capability grants several others:
\begin{enumerate}
    \item upstream inference and classification of molecular identities;
    \item sensor-signature classification to enable tighter robotic hardware integration and counter drift effects;
    \item retrieval of aroma descriptors correlating with various molecules;
    \item mapping and ranking of image regions to localize odor profiles;
    \item physical source localization of an odor that can be extrapolated from a 2D image into 3D space.
\end{enumerate}


The abundance of visual and linguistic data online creates convenient avenues on which robust trillion-parameter vision-language models (VLMs) may be trained.
However, an olfactory analog at this scale does note  \cite{Munger-parma-2025-chemosensory-barriers, PARMA2026105545}.
We believe there is an inherent limit on what machines can perceive in the chemical world around them due to the lack of standardized olfactory data that can be used for constructing robust world models with detailed chemistry.

Many strong VLMs already align hundreds of millions of image-text pairs \cite{radford2021clip, zhai2023siglip}.
Within these VLMs, many of these words are odor descriptors. 
As a result, odor descriptors can be used as a bridge between chemistry and language: descriptor words such as ``citrus'', ``smoky'', and ``floral'' already live in image-caption space, so an initial olfaction-vision grounding can be established through shared geometry without joint training  \cite{marki_sisson_satarifard_2026benchmarkassessingolfactoryperception}.
Thus, the model already has momentum toward implicitly placing aromas near the images they describe without arduous data collection.

Our intent with \textit{COLIP-2} is to demonstrate the limits of what vision-olfactory perception can achieve when trained on open - yet disparate - olfactory datasets and web-scale visual-linguistic knowledge that feeds many of the VLMs with which we interact with every day.
\textit{COLIP-2} demonstrates compelling performance on many measures, but it is not perfect in its own right.
This emphasizes the need for more progress in three domains, specifically for olfaction models: (1) the collection of more data oriented toward robotics, (2) the establishment of more olfactory benchmarks, and (3) protocols for continual learning based on minimal data.
\textit{COLIP-2} sets up the track for progress to be made in all three.

\textit{COLIP-2} trains additional modality encoders against a vision-language backbone to create a new shared anchor space in the style of \textit{ImageBind} \cite{girdhar2023imagebind}. 
Olfaction-vision grounding then arises through the shared text geometry, with no paired scent-image data required.
This variant of \textit{COLIP} emphasizes what is possible on existing datasets without additional data collection on elaborate lab instrumentation.\footnote{The most robustly grounded dataset would consist of an entirely new set of samples where olfactory sensors, cameras, and a trained panel of aroma experts work together to align all modalities. No such data currently exists. Scentience is establishing this dataset now and will use the embeddings created here to complement alignment in a future model release.}

One mistake we made early on in development was to treat all chemical sensors as equal.
Many chemical sensors exist on the market today, and each outputs a different value or series of values.
Arguably none of these sensors are optimized for fast inference for real-time robotics applications such as scent-based navigation.
We take this into consideration by creating a custom encoder to normalize sensor readings from the most common third-party sensors (as well as our own custom sensors) to alleviate one more barrier to entry for the robotics community.
Table \ref{tab:details} enumerate some core details of the model.

\begin{table}[h]
\caption{Model Details}
\label{tab:details}
\vspace{2mm}
\centering\small\sans
\begin{tabular}{L{3.0cm} L{12.9cm}}
\hdrow \thd{Field} & \thd{Value} \\
Model name        & Scentience COLIP-2 Embeddings \\
\rowcolor{scrowalt} Version / date & v2.0.1 \textperiodcentered\ July 2026 \\
Task              & Olfaction-vision-sensor grounding via a shared embedding space \\
\rowcolor{scrowalt} Status & Robotics research prototype; core alignment complete and evaluated \\
Modalities        & Molecular structure, gas-sensor reading, odor-descriptor text, image \\
\rowcolor{scrowalt} Backbones & Swappable VLM core: high-fidelity server or compact on-device variants \\
Shared space      & One embedding space shared across all four modalities \\
\rowcolor{scrowalt} Intended use & Research prototype for robotics applications; not validated for safety-critical use (see \S\ref{sec:limitations},~\S\ref{sec:license}) \\
\end{tabular}
\end{table}

At a conceptual level the system has two encoders that all emit a comparable vector in the same space, plus a frozen backbone that supplies the anchors:

\begin{itemize}[leftmargin=1.2em,itemsep=2pt,topsep=2pt]
  \item A \textbf{trainable olfactory (molecule) encoder}: a compact graph transformer that reads
        a molecule's structure (atoms and their connectivity) and produces an odor embedding.
        Two model variants are constructed: (1) a larger, more accurate model intended for cloud-based serving, and (2) a smaller and slightly less accurate model that is deliberately built from operations that export cleanly to on-device runtimes.
        We focus on the latter with this model card because our primary motive is to demonstrate a solution optimized for edge-based AI and robotics.
  \item A \textbf{trained chemical sensor encoder}: a small time-series model that turns various e-nose sensor readings into an embedding in the same space.
  This model is agnostic to the sensing substrate, including those of the most com: metal-oxide, electrochemical, non-dispersive infrared, spectrometry, photo-ionization, acoustic, and others.
  \item A \textbf{vision-language backbone}: the image and text encoders of the VLM are minutely fine-tuned and then fixed such that a controlled vocabulary of odor descriptors is encoded once to serve as the shared
        anchors to which every other modality is aligned.
\end{itemize}

Once molecule, sensor, descriptor-text, and image all land in one space, any modality can query any
other by similarity, and the odor-source question becomes a ranking of image regions against a
detected-aroma vector.

\begin{figure}[h]
\centering
\begin{tikzpicture}[
    font=\sffamily\small,
    box/.style={rounded corners=4pt, draw=scnavy!55, line width=0.7pt, align=center,
                inner sep=5pt, minimum height=9mm, text=scnavy},
    train/.style={box, fill=green!8, draw=green!45!black!55},
    frozen/.style={box, fill=orange!12, draw=orange!55!black!45},
    infer/.style={box, fill=red!7, draw=red!45!black!45},
    shared/.style={rounded corners=4pt, draw=scnavy, line width=1.1pt, fill=scnavy!8,
                   align=center, inner sep=6pt, text=scnavy, font=\sffamily\bfseries\small},
    grp/.style={rounded corners=6pt, draw=scgray!60, dashed, line width=0.6pt, inner sep=8pt},
    lbl/.style={font=\sffamily\footnotesize\itshape, text=scgray},
    a/.style={-{Stealth[length=2mm]}, draw=scnavy!70, line width=0.7pt},
    ad/.style={-{Stealth[length=2mm]}, draw=scgray!80, line width=0.7pt, dashed},
  ]

  \node[train] (mol)  {Molecule encoder\\{\footnotesize\upshape (structure $\rightarrow$ odor)}};
  \node[train, below=6mm of mol] (sen) {Sensor encoder\\{\footnotesize\upshape (sensor $\rightarrow$ odor)}};
  \begin{scope}[on background layer]
    \node[grp, fit=(mol)(sen), label={[lbl]above:olfaction encoders}] (trainbox) {};
  \end{scope}

  \node[shared, right=20mm of $(mol)!0.5!(sen)$] (space) {Shared\\embedding\\space};

  \node[frozen, above right=6mm and 16mm of space] (anchor) {Descriptor anchors\\{\footnotesize\upshape (text encoder)}};
  \node[frozen, right=10mm of anchor] (img) {Image encoder};
  \begin{scope}[on background layer]
    \node[grp, fit=(anchor)(img), label={[lbl]above: vision-language model}] (frozenbox) {};
  \end{scope}

  \node[infer, below right=7mm and 16mm of space] (cam) {Camera regions};
  \node[infer, right=8mm of cam] (match) {Cosine match\\{\footnotesize\upshape rank objects by aroma}};
  \begin{scope}[on background layer]
    \node[grp, fit=(cam)(match), label={[lbl]below:inference: ``which object in the scene emits this aroma?''}] (inferbox) {};
  \end{scope}

  \draw[a] (mol.east) -- (space.west);
  \draw[a] (sen.east) -- (space.west);
\draw[a] (anchor.south) -- node[lbl,right=1pt]{} (space.north east);
  \draw[ad] (img.south) to[out=-90,in=20] (space.east);
  \draw[a] (space.south east) -- (cam.north west);
  \draw[a] (cam.east) -- (match.west);
  \draw[ad] (space.east) to[out=-30,in=150] (match.north west);

\end{tikzpicture}
\caption{\small Trainable olfactory and sensor encoders (green) are aligned
into the shared space of a vision--language backbone (amber) using descriptor anchors
as targets. At inference, camera regions and a detected aroma are embedded into the same space and
matched by similarity to answer ``which object emits this scent?''.}
\label{fig:arch}
\end{figure}
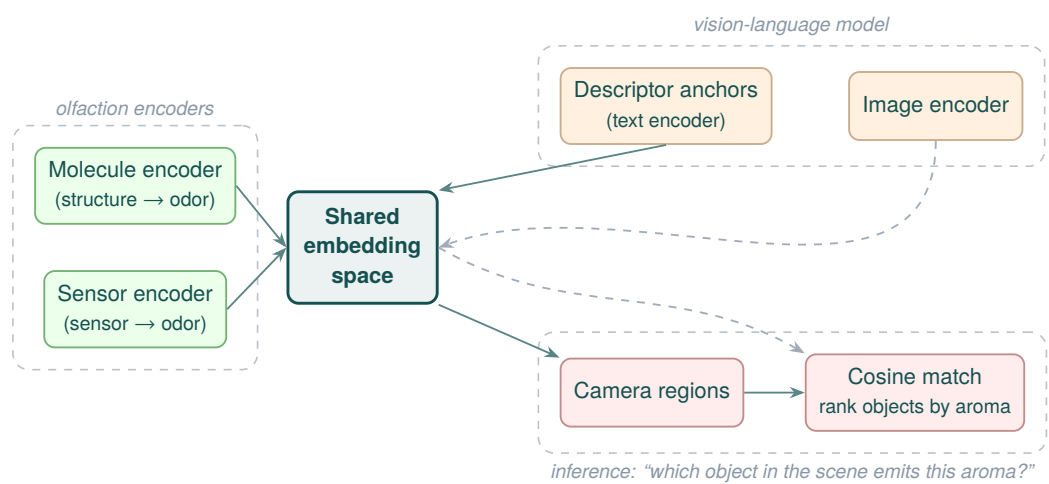
\section{Improvements from \textit{COLIP-1}}

Scentience \textit{COLIP-2} is the second-generation successor to \emph{COLIP-1} \cite{france2025diffusiongraphneuralnetworks,france2025colip-ovlembeddings}, an earlier Scentience model that added an olfactory encoder to \textit{CLIP} and trained the two \emph{jointly} with a symmetric contrastive objective (both the olfactory encoder and the vision) language encoders moving together. 
\textit{COLIP-2} maintains the thesis of binding olfaction into a vision-language shared space via contrastive objective, but with some architectural adjustments along with the addition of a gas-sensor encoder.
The vision-language backbone and olfactory and gas-sensor encoders are trained and aligned one-sidedly into the backbone's descriptor anchors. 
Symmetric co-training can only ground scent to vision where paired scent-image data exists (which it does not at scale) so building on a solid vision-language backbone and aligning through the descriptor bridge is what lets olfaction$-$image grounding \emph{emerge} with no scent-image pairs. 
\textit{COLIP-2} is therefore best read as a re-design and re-grounding of \textit{COLIP-1}, versus a strict re-train \textit{COLIP-1}. 

One other major difference between \textit{COLIP} versions is in the training data.
Synthetic training data was specifically generated for \textit{COLIP-1}.
For \textit{COLIP-2}, we took a different route after receiving constructive feedback from the sensory sciences, robotics, chemistry, and AI communities.
By utilizing a higher-caliber VLM and leveraging the aroma relationships already built into the model, \textit{COLIP-2} inherits web-scale vision–language knowledge at a fraction of the training and storage cost.
This relieves the need to synthetically construct olfaction-vision relationships and provide another variable that could corrupt grounding.
As a result, olfaction-image grounding emerges through shared geometry without paired smell–image data.

Finally, we solicited honest feedback on \textit{COLIP-1} from scientists across disciplines in both academia and industry, including those from perfumery, neuroscience, food science, biochemistry, robotics, AI, and more.
This feedback was reconciled in architectural changes and a complete re-write of the training pipeline, along with other improvements noted throughout this document.
We acknowledge that moving olfaction forward for AI and robotics requires intimate cross-collaboration among leading researchers in a variety of industries beyond AI and robotics.
Consequently, while \textit{COLIP-2} is nowhere close to multimodal model engineering limits, we feel it is much more aligned with human olfaction.

\section{Model specification}
The vision-language backbone and olfactory encoders are both configurable and swappable.
We construct two options to support two different operating points:

\begin{enumerate}
    \item \textbf{COLIP-2 Cloud:} a high-fidelity server model that boasts maximum accuracy and high-resolution heatmaps intended for serving via cloud endpoint. The olfactory encoder is based on a graph-attention transformer.
    \item  \textbf{COLIP-2 Edge:} a smaller \textit{int8} quantized model optimized for edge-based inference on CPU or small GPUs, enabling a real-time, offline aroma heatmap. Like its cloud counterpart, the olfactory encoder is also graph-based, but with ops optimized for hardware-agnostic export.
\end{enumerate}

\begin{figure}[t]
\centering
\includegraphics[width=0.4\linewidth]{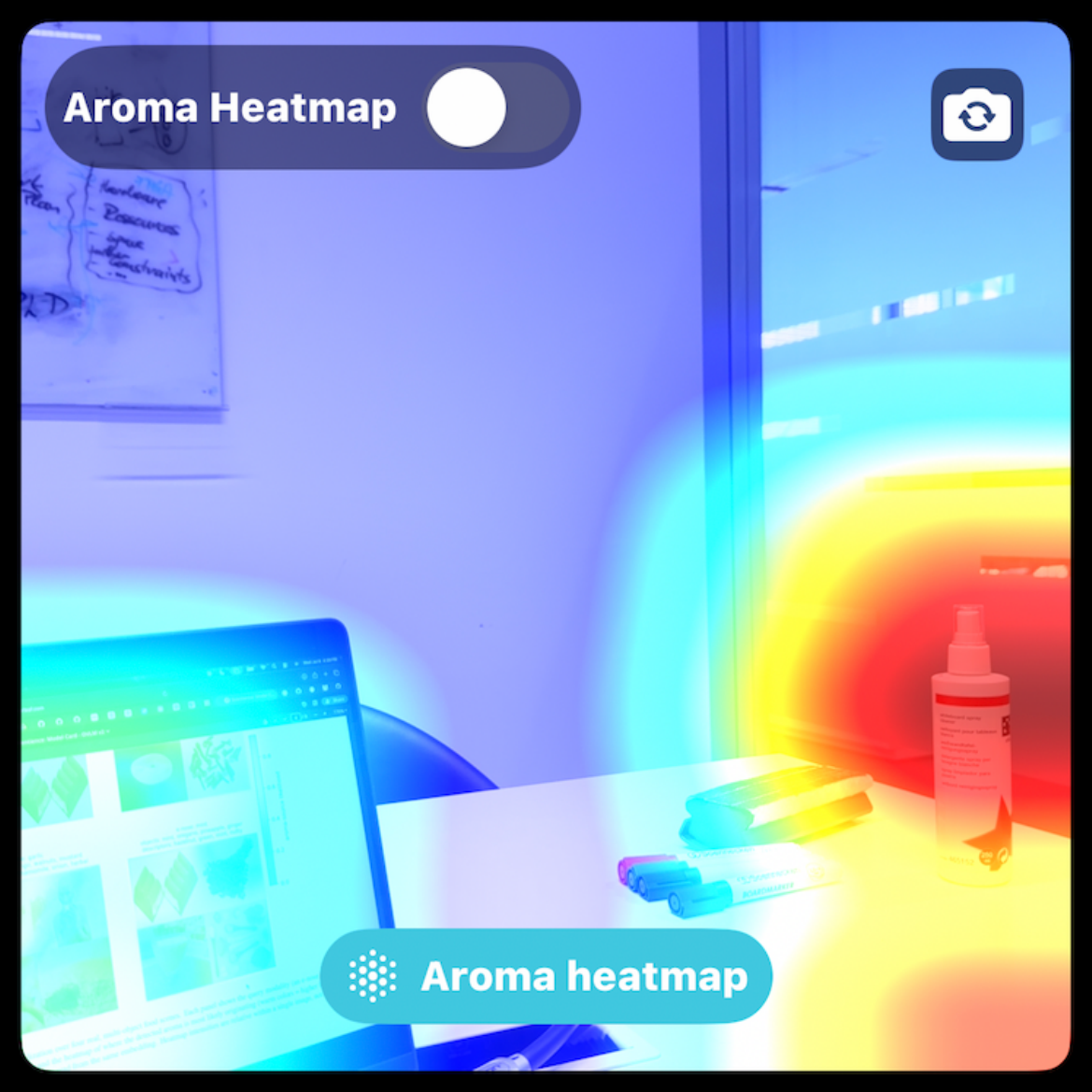}
\caption{\small Aroma localization for ethanol via \textit{COLIP-2}.}
\label{fig:heatmap_ios}
\end{figure}

The language-side and substance-identification tasks are essentially backbone-invariant; the main cost of the backbone lands on exact image grounding, which is the intended trade for real-time on-device inference and high-resolution heat-mapping. 
An example of this can be observed in the \textit{COLIP-2} snapshot from the Scentience iOS application in Figure \ref{fig:heatmap_ios}.

The trainable olfactory and sensor encoders are small and export to an on-device runtime, so molecule and sensor embedding can run fully offline. 
Each Scentience instrument runs this encoder on-device and continuously updates itself via federated learning (see Section \ref{sec:federated_learning} for details).
Access to \textit{COLIP-2 Cloud} is offered through any of the hosted APIs via API key.
\textit{COLIP-2 Edge} is offered within the Scentience iOS and Android mobile applications \cite{scentience2025_ovlm}.

Unlike a large language model, \textit{COLIP-2} is not compressed to low-bit weight formats and there is nothing large to quantize. 
All encoders are trained in FP32 and exported to FP16 for on-device inference.
The number of trainable parameters sums to 1.15B for \textit{COLIP-2 Cloud} and 101M for \textit{COLIP-2 Edge}.
Further details on parameter counts and storage footprints can be observed in Table \ref{tab:footprint}

\begin{table}[h]
\label{tab:footprint}
\centering\small\sans
\caption{\small Storage footprint. Native = untouched FP32; on-device = FP16. Sizes are decimal.}
\begin{tabular}{L{4.6cm} L{3.5cm} L{3.5cm} L{2.9cm}}
\hdrow \thd{Component} & \thd{Trainable Params} & \thd{Cloud Footprint} & \thd{On-Device Footprint} \\
\textit{COLIP-2} Cloud & 1.15\,B & 4.4\,GB & --- \\
\rowcolor{scrowalt} \textit{COLIP-2} Edge & 101\,M & 386\,MB & 72\,MB \\
Olfactory encoder (isolated) & 1.30\,M & 5.0\,MB & 1.2\,MB \\
\rowcolor{scrowalt} Sensor encoder (isolated) & 0.63\,M & 2.4\,MB & 1.2\,MB \\
Descriptor anchors ($175 \times D$) & --- & 0.8\,MB & 0.4\,MB \\
\end{tabular}
\end{table}

\section{Intended use \& scope}
\textit{COLIP-2} is intended to be used for robotics research, such as in scent-based navigation augmented by vision.
It is also designed to complement existing AI foundation models needing cross-modal bi-directional scent$-$vision retrieval.
However, any application requiring a mapping of
detected molecules to candidate source objects in an image, or zero-shot prediction of a molecule's odor-descriptor profile, is an excellent candidate of use.
Of particular interest is the aroma heatmap feature that is enabled by the model, an example of which can be observed in Figure \ref{fig:heatmap_ios}.
\textit{COLIP-2} materializes real-time chemical localization over an image by creating an RGB heatmap around the ``hottest'' objects from which an aroma may be coming.
Each molecule or aroma produces its own distinct heatmap.

The model requires the use of olfaction sensors and cameras, so it assumes the user has raw olfactory output and an image for input.
The sensor encoder is design to encode molecular data into a universal format that can be interpreted by the olfactory encoder.
With the most common third-party chemical sensors are supported, this means that the users does not have to add in any extra sensor-dependent harnessing to achieve results.
While the normalization of common olfactory sensors has been accommodated via the sensor encoder, \textit{COLIP-2} does not account for any drift or environmental effects encountered by the sensor, nor does it account for any other sources of bias within the sensor readings.
The olfaction sensor is assumed to yield a molecular identity for input into the \textit{COLIP-2}. 
Any additional sensor$-$molecule front-end required by the sensor manufacturer, which is not accommodated by this model.

\textit{COLIP-2} can be accessed via the Scentience APIs \cite{scentience2025_ovlm, scentience-pypi, scentience-crate, scentience-npm, scentience-conan}.
The sensor encoder is designed to support common third-party sensor families through device-specific normalization and calibration. 
Performance should be validated independently for each sensing platform to ensure proper functionality.
The encoder has also been trained to work with Scentience olfactory hardware;
for access to a Scentience instrument, please contact \href{mailto:info@scentience.ai}{info@scentience.ai}. 

\textit{COLIP-2} is not intended for nor evaluated on safety-critical applications pertaining to toxic-exposure alarms or food-safety decisions.
\textit{COLIP-2} is also not designed for healthcare applications and can not contribute to any medical or diagnostic claims.
We discourage the use of \textit{COLIP-2} in any setting where a missed or wrong detection can cause harm. 
\section{Capabilities}
Because molecule, sensor, aroma text, and image all share one space, any modality can query the other by similarity. 
These multi-dimensional relationships enable powerful queries, but have some limitations that prevent them from being a silver bullet.
Table \ref{tab:capabilities} below summarizes what is exposed today, a reasonable expectation of the relationship strength, and how we are currently laying the groundwork for the next version.

\begin{table}[h]
\caption{Model Capabilities}
\label{tab:capabilities}
\vspace{2mm}
\centering\small\sans
\begin{tabular}{L{2.4cm} L{2.8cm} L{4.0cm} L{2.6cm} L{3.2cm}}
\hdrow \thd{Query\(\downarrow\)/retrieve\(\rightarrow\)}  & \thd{Odor descriptors} & \thd{Scene image (grounding)} & \thd{Substance / class} & \thd{Path Forward} \\
Molecule     & Zero-shot odor profile        & Weak (single-compound proxy)      & N/A & Gather instrumented lab data \\
\rowcolor{scrowalt} Sensor reading & Strong & Family-level grounding & Strong substance ID & Improve sensor design \& modeling \\
Aroma descriptor   & Strong                        & Clean deployment path             & Strong & Curate panelist olfactory data \\
\rowcolor{scrowalt} Image region & Reads off the aroma & Region $\leftrightarrow$ region & N/A & Gather multimodal world data \\
\end{tabular}
\end{table}

The model is strongest at the odor-family level throughout. 
Exact-instance identification and recognition of genuinely novel substances are known weak paths (see \S\ref{sec:limitations}).
``N/A'' marks pairs that do not form a meaningful query.

\section{Training data}
The \textit{COLIP-2} training repository is curated from a diverse collection of license-permitting data encompassing a wide range of olfactory domains and publicly available vision-language datasets.
We augment this curated training set with additional public data from third party olfactory sensors that comprise of electrochemical, optical, acoustic, metal oxide, and other sensing mediums.
Since no single chemical sensor enables superior performance in all domains, we elect to accommodate as many sensors as possible.
All data is normalized into a single controlled odor-descriptor ontology and keyed by molecular identity in order to devise a unified training schema. 





\section{Evaluation}\label{sec:eval}
Evaluation metrics for \textit{COLIP-2} can be observed in Table \ref{tab:eval_metrics}.
All numbers are measured on held-out data unseen during training. 
Odor-descriptor retrieval ranks a controlled set of descriptor anchors by similarity to a molecule's embedding; a descriptor is counted relevant when its soft applicability label is at least 0.5.

\begin{table}[h]
\caption{Evaluation Metrics}
\label{tab:eval_metrics}
\vspace{2mm}
\centering\small\sans
\begin{tabular}{L{9.5cm} R{3.0cm}}
\hdrow \thd{Metric} & \thd{Value} \\
Odor-descriptor AUROC (chemical grounding retained)    & 0.84 \\
\rowcolor{scrowalt} Olfaction$-$descriptor Recall@1 & 0.62 \\
Olfaction$-$descriptor Recall@5               & 0.90 \\
\rowcolor{scrowalt} Olfaction$-$descriptor Recall@10 & 0.94 \\
Gas-sensor $-$ source substance, top-1        & 0.90 \\
\rowcolor{scrowalt} Gas-sensor $-$ source substance, Recall@5 & 0.96 \\
Gas-sensor $-$ scene image, exact top-1       & 0.72 \\
\rowcolor{scrowalt} Gas-sensor $-$ scene image, category top-1 & 0.90 \\
\end{tabular}
\label{tab:performance}
\end{table}

For roughly 90\% of held-out molecules, a correct odor descriptor appears in the top five retrieved.
Because those descriptors are already aligned to images by the VLM backbone, a never-before-seen molecule lands near the correct visual concepts.
The fact that the descriptor AUROC is unchanged after alignment confirms the binding did not cost chemical accuracy. 
The sensor encoder enabled real e-nose readings to identify their source substance at 0.90 top-1 and, on held-out recordings, to land on the correct object in a scene at the odor-family level.

\section{Aroma-source localization}
Beyond identifying \emph{which} object emits a detected aroma, the system can show from \emph{where} in an image the aroma is coming, as a color-coded overlay. 
This enables new capabilities for not only robots, but mobile devices as well - point a camera at a scene, detect a scent, and highlight the most probablistic sources. 
The query aroma can come from any of the three input modalities: a molecule, a live e-nose reading, or descriptor text.
Since all three modalities share the same space, the same query also reads off the top odor descriptors.

\begin{figure}[t]
\centering
\includegraphics[width=0.82\linewidth]{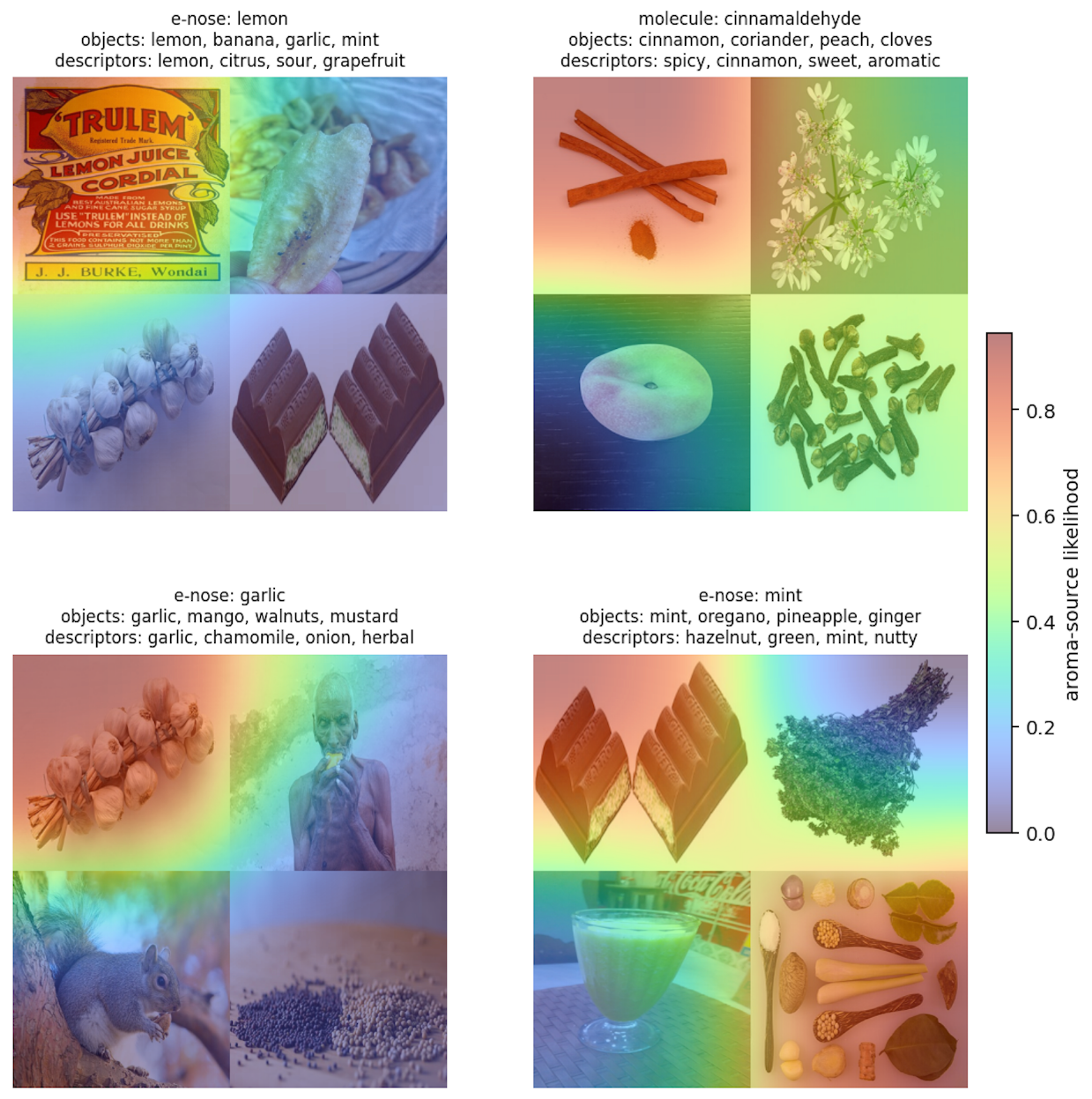}
\caption{\small Aroma-source localization over four real, multi-object food scenes. Each panel shows the query modality (an e-nose reading or a molecule), the objects in view, and the heatmap of where the detected aroma is most likely originating (warm colors = higher likelihood), together with the top odor descriptors read from the same embedding. Heatmap intensities are relative within a single image, not calibrated probabilities.}
\label{fig:heatmap}
\end{figure}

This feature itself enables many capabilities.
Within the Scentience APIs are additional logic to amplify the effects of this model feature for robotic navigation.
For example, if a robot is tasked with navigating to the source of an odor by tracking a target molecule, the molecule may evolve through interactions with different molecules in the air.
Odor mapping and navigation techniques are built in to contextualize several heatmaps together in time series and develop a live plume model.
Additional checks for hysteresis and drift are also integrated to ensure the heatmap is aligned with the sensing hardware to foster the highest accuracy possible.

\begin{figure}[h]
\centering
\includegraphics[width=1.0\linewidth]{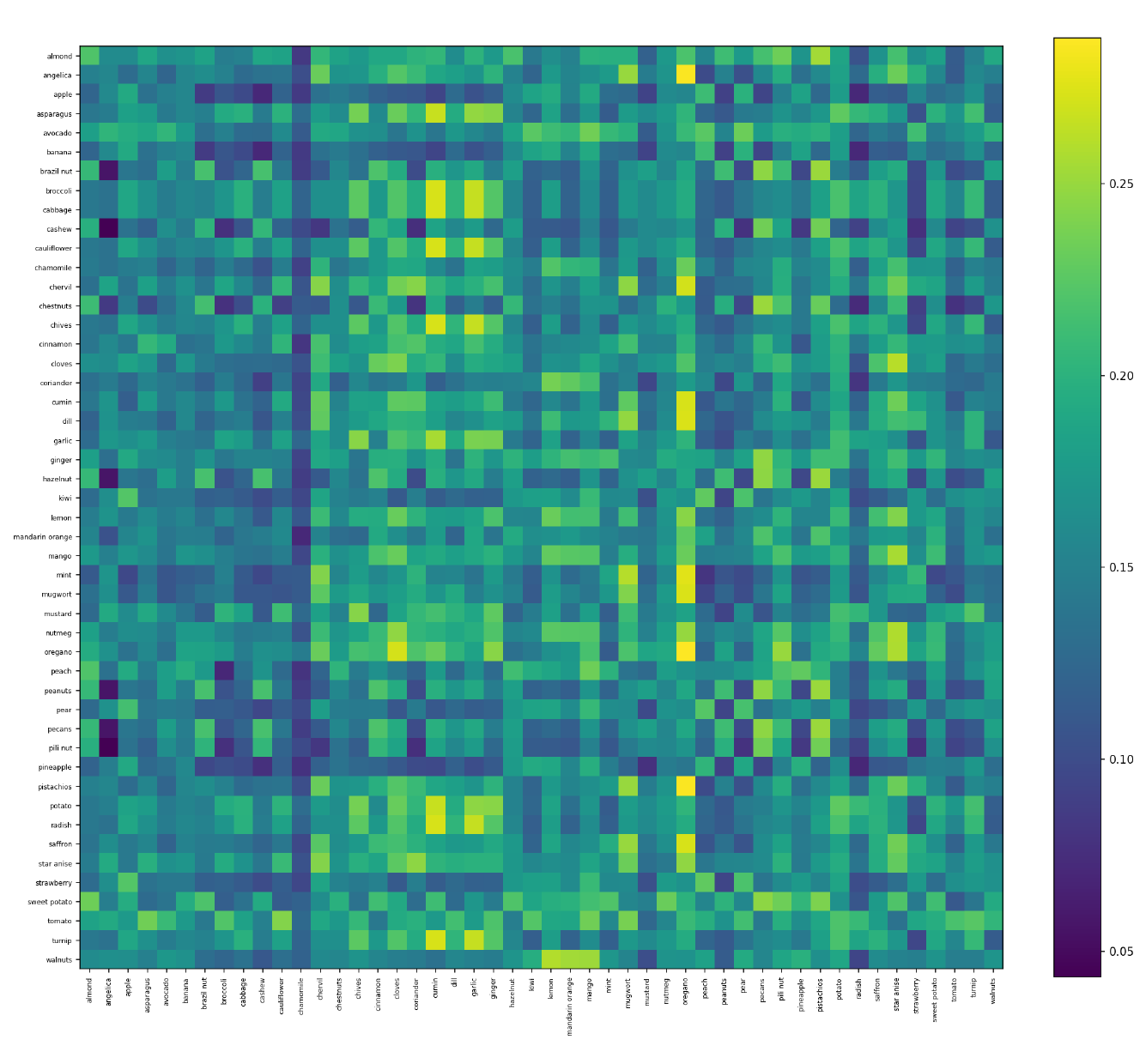}
\caption{\small Aroma-image similarity across the held-out food gallery (rows = aroma
query, columns = food image). A bright diagonal is correct grounding; the off-diagonal bands are the
odor-adjacent confusions above. The structure is strongest at the odor-family level.}
\label{fig:confusion}
\end{figure}

Figure \ref{fig:heatmap} demonstrates a panel of heatmaps for aroma-source localization, while Figure \ref{fig:confusion} demonstrates aroma-image similarity across a hold-out set of food-like images.

%
%
%

\section{Finetuning Ablation Study}\label{sec:embgeom}
We hypothesize that useful olfaction-vision relationships can be interpolated by binding molecular and sensor encoders into the latent space of a trained vision-language model.
Because the text encoder of the VLM is co-located with the image encoder, an aroma that retrieves the descriptor ``smoky'' also lands near smoky-looking food images.
Thus, light grounding emerges with no odor-image training pairs. 
However, we desire to improve upon this grounding and strengthen the odor-image relationships that exist.
The central risk of doing so is that finetuning the text encoder of the VLM to sharpen olfaction descriptor retrieval could quietly drag it off the image geometry and silently break grounding. 
We conduct a series of probing experiments to measure exactly that, holding everything except the backbone fixed.
Then, as a final boundary condition, we deliberately remove every safety rail to show what a genuine break in embedding alignment looks like.

In light of the above, we perform an ablation over five backbone conditions: one frozen, three controlled LoRA finetunes, and one deliberately over-finetuned boundary condition designed to find the breaking point of olfactory-vision relationships.
During each ablation, the trained olfactory encoder is held frozen as a constant to deduce whether finetuning the vision-language backbone strengthens or weakens the olfaction-vision and olfaction-language relationships. 

\subsection*{Experimental design}
Three anchored entity types share one joint space (all mean-centered and L2-normalized so they are directly comparable) observable in Table \ref{tab:encoder_roles}

\begin{table}[h]
\caption{Encoder Roles}
\label{tab:encoder_roles}
\vspace{2mm}
\centering\small\sans
\begin{tabular}{L{4.5cm} L{4.1cm} L{3.2cm} L{3.3cm}}
\hdrow \thd{Entity} & \thd{Produced by} & \thd{Role} & \thd{State across conditions} \\
\textbf{Olfaction}: 610 held-out molecules & olfactory encoder & aroma capture & frozen \\
\rowcolor{scrowalt} \textbf{Language}: 175 aroma anchors & VLM text encoder & vision-olfaction bridge & variable \\
\textbf{Vision}: 48 food-like images & VLM image encoder & image capture / camera & frozen \\
\end{tabular}
\end{table}

The olfactory encoder is frozen for all five configurations, as is the
image encoder as a means of control: if only the text embeddings can move, any change
in the cross-modal geometry is attributable to the VLM backbone alone. 
Additionally, freezing the aligned encoder, caching its embeddings once, and taking a full-dataset gradient matches our deployment recipe.
Freezing the olfactory embeddings also allows us to answer the question, ``if I move the backbone under a fixed olfaction
manifold, does the cross-modal relationship get better or worse?'' 
Co-adapting the olfactory encoder could be an enlightening experimental battery in itself, and we note this as a prospect for future work.

We designate the five different studies as \textit{configurations 0-4}, formally designated as \textit{C0-C4} for brevity.
Each configuration can be observed in Table \ref{tab:ablation_configs}.
C0 denotes the configuration associated with the frozen VLM.
C1--C3 form a controlled sweep: we hold the optimizer, learning rate
($3\times10^{-4}$), 18-step budget, gradient clip (1.0), and seed constant leaving only two tuning knobs.
Both tuning parameters trade language-alignment against vision-grounding: (1) the adapter capacity (LoRA rank + which projections it adapts), and (2) a defined variable $\lambda$ that represents the anchor-distillation guardrail strength. 
C4 denotes the boundary condition configuration and we deliberately hold it outside that sweep; it removes the guardrail and the safety of a gentle schedule with a fully tunable text encoder, a more aggressive learning rate, and is trained until it collapses.

\begin{table}[h]
\caption{Configurations for Ablation Study}
\label{tab:ablation_configs}
\vspace{2mm}
\centering\footnotesize\sans
\begin{tabular}{L{0.5cm} L{2.5cm} L{1.15cm} L{2.55cm} R{1.2cm} L{0.8cm} L{2.85cm}}
\hdrow \thd{ID} & \thd{Name} & \thd{Rank\,/\,$\alpha$} & \thd{Target modules} & \thd{Params} & \thd{$\lambda$} & \thd{Schedule} \\
\textbf{C0} & Frozen (off-the-shelf) & --- & --- & 0 & --- & --- \\
\rowcolor{scrowalt} \textbf{C1} & Guarded LoRA & 8\,/\,16 & \texttt{q,v} & 995\,K & 1.0 & lr 3e-4, clip 1, 18\,st \\
\textbf{C2} & Balanced LoRA & 16\,/\,32 & \texttt{q,v} & 1.99\,M & 0.3 & lr 3e-4, clip 1, 18\,st \\
\rowcolor{scrowalt} \textbf{C3} & Unconstrained LoRA & 16\,/\,32 & \texttt{q,k,v,out} & 3.98\,M & 0.0 & lr 3e-4, clip 1, 18\,st \\
\textbf{C4} & Over-finetuned (guardrail off) & 32\,/\,64 & \texttt{q,k,v,out,fc1,fc2} & 17.4\,M & 0.0 & \textbf{lr 4e-3, clip 12, 40\,st} \\
\end{tabular}
\end{table}

A low-rank adapter on the attention projections is the least-destructive way to move the text geometry. 
C1/C2 adapt only query and value projections; C3 adds key + output; C4 additionally adapts the MLP layers at rank 32 for maximum disruption.
The $\lambda$ variable acts as a critical throttle for anchor alignment.
It is a guardrail constraining the movement of text anchors to their image-aligned positions.
$\lambda=1.0$ (C1) holds them almost still; $\lambda=0.3$ (C2) lets them relax; $\lambda=0$ (C3, C4) removes the guardrail.

At matched learning rate and unit gradient clip, the LoRA adapter capacity and $\lambda$ guardrail are the only explanations for any disparity between C1--C3. 
C4 uses $13\times$ the learning rate and a $12\times$ looser clip: the clip in C1--C3 caps how far each step can move the geometry, so raising it is what lets C4 escape the alignment objective's stabilizing pull and actually break the text encoder.
We define ``anchor drift'' as the mean cosine of each finetuned anchor to its frozen position.
    

\subsection*{Results}

\begin{figure}[t]
\centering
\includegraphics[width=\linewidth]{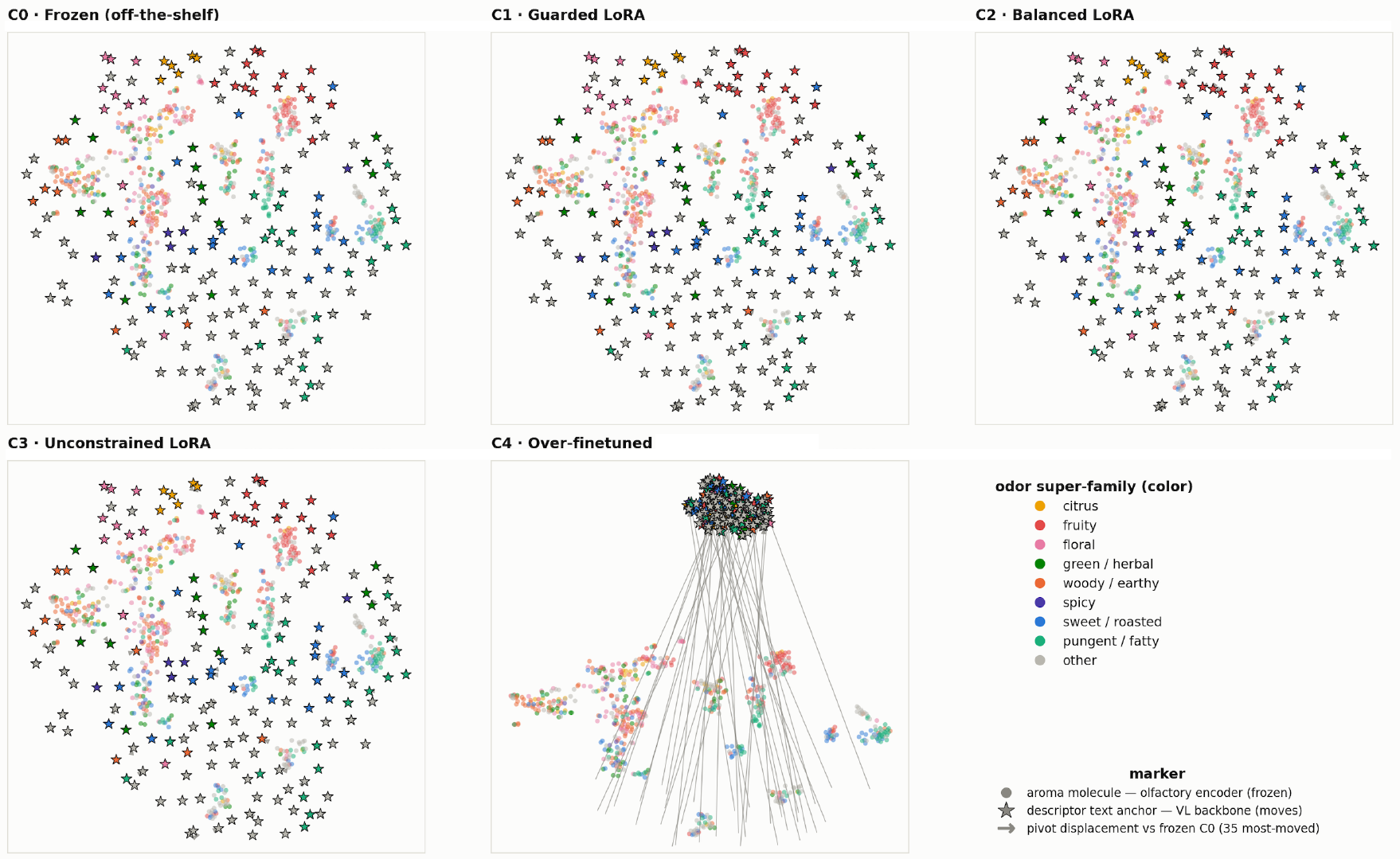}
\caption{\small t-SNE plots for each configuration: The trained olfactory encoder's 610 held-out molecule embeddings (dots, colored by odor super-family) together with the backbone's 175 descriptor-text anchors (stars), in the shared text space. A single global t-SNE, so the aroma cloud is identical in every panel and only the
anchors move; arrows mark the 35 most-displaced anchors relative to the frozen (C0) geometry. C0-C3 keep the
anchors interleaved with the aroma clusters; C4 collapses all 175 anchors into a single clump torn
completely away from the olfaction cloud. Note: food images live in a separate modality cone, so they are reported through the grounding metrics rather than plotted.}
\label{fig:embgeom-tsne}
\end{figure}

\begin{table}[h]
\caption{Finetuning Ablation Metrics}
\label{tab:ablation}
\vspace{2mm}
\centering\small\sans
\begin{tabular}{L{5.6cm} R{1.7cm} R{1.7cm} R{1.7cm} R{1.7cm} R{1.9cm}}
\hdrow \thd{Metric (higher is better)} & \thd{C0} & \thd{C1} & \thd{C2} & \thd{C3} & \thd{C4} \\
Olfaction$-$language R@1        & 0.627 & 0.628 & 0.620 & 0.618 & \textbf{0.025} \\
\rowcolor{scrowalt} Olfaction$-$language R@5 & 0.878 & 0.885 & 0.883 & 0.885 & \textbf{0.176} \\
Olfaction$-$language R@10       & 0.911 & 0.921 & 0.926 & 0.934 & \textbf{0.293} \\
\rowcolor{scrowalt} Olfaction$-$language mAP & 0.449 & 0.463 & 0.471 & 0.479 & \textbf{0.062} \\
Anchor drift (cos to frozen)              & 1.000 & 0.991 & 0.969 & 0.791 & \textbf{0.005} \\
\rowcolor{scrowalt} Olfaction$-$vision category top-1 & 0.542 & 0.542 & 0.521 & 0.542 & \textbf{0.167} \\
Olfaction$\-$vision descriptor R@5 & 0.271 & 0.271 & 0.271 & 0.271 & \textbf{0.104} \\
\rowcolor{scrowalt} Text$-$image ceiling, category top-1 & 0.938 & 0.938 & 0.938 & 0.938 & \textbf{0.062} \\
Direct (z vs image) category top-1 \emph{(control)} & 0.583 & 0.583 & 0.583 & 0.583 & \textbf{0.583} \\
\end{tabular}
\end{table}

\begin{figure}[t]
\centering
\includegraphics[width=0.96\linewidth]{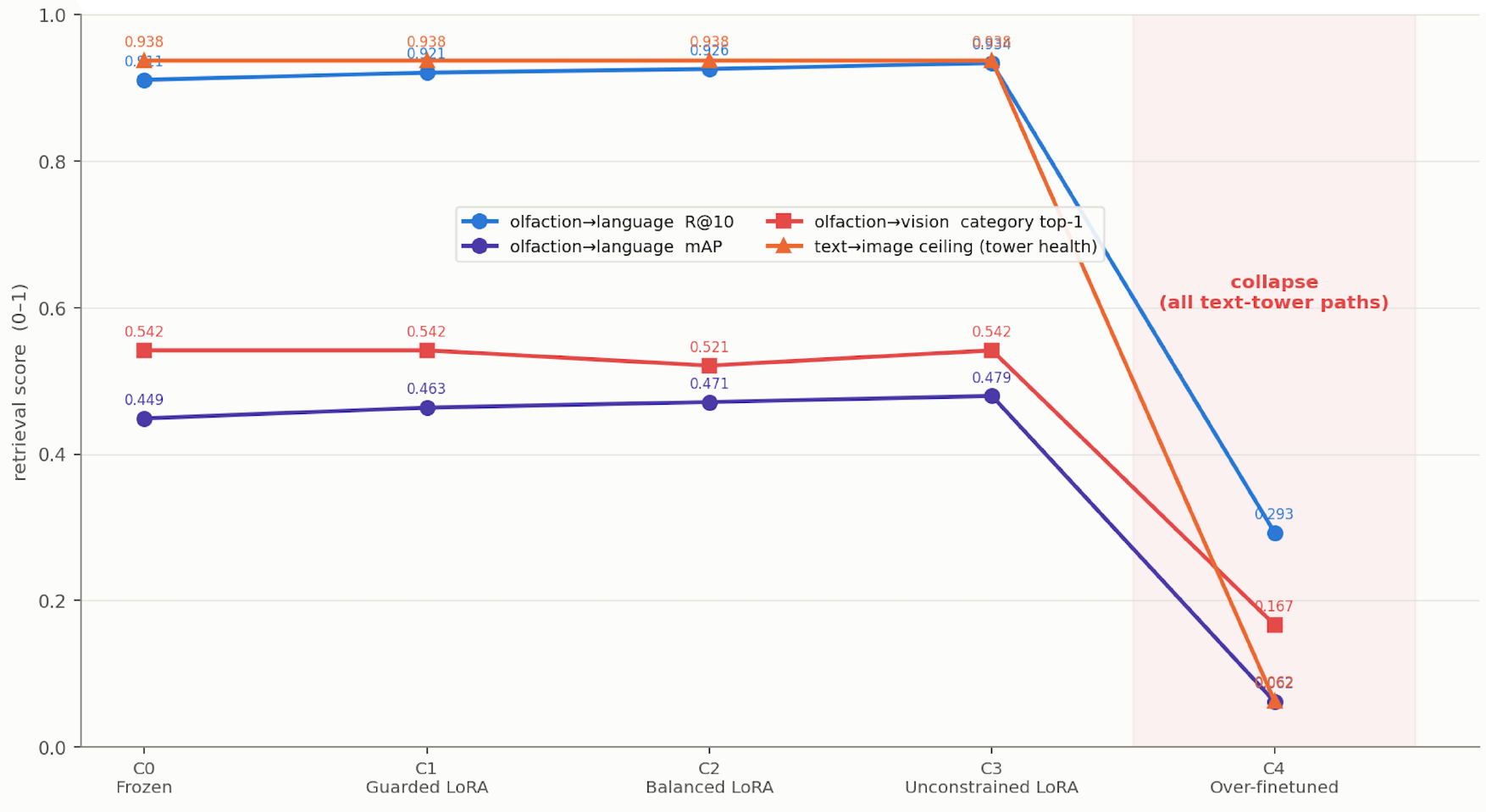}\\[4pt]
\includegraphics[width=0.96\linewidth]{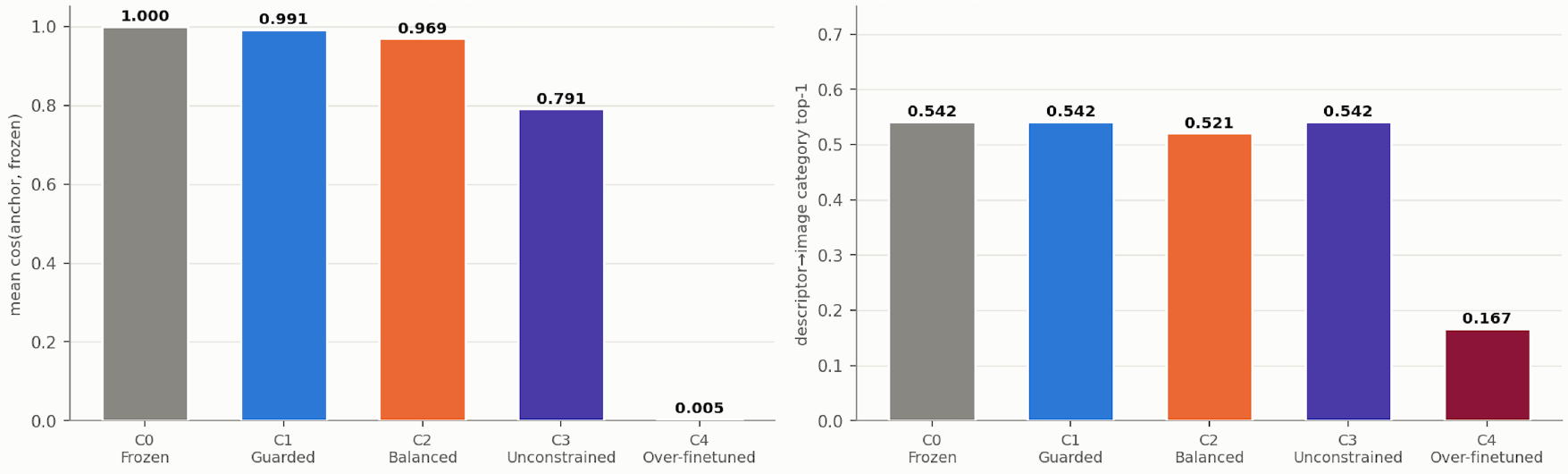}
\caption{\small \textbf{Top:} across C0-C3 the olfaction-language metrics rise while olfaction-vision grounding and the text-image ceiling stay flat; at the over-finetune (C4) all four rapidly diverge together. \textbf{Bottom:} the guardrail dials anchor drift smoothly ($1.00\to0.99\to0.97\to0.79$) while grounding holds, until C4 drives drift to $0.005$ (left) and grounding to the random baseline, going from $0.542$ to $0.167$ (right).}
\label{fig:embgeom-summary}
\end{figure}

\begin{figure}[h]
\centering
\includegraphics[width=0.94\linewidth]{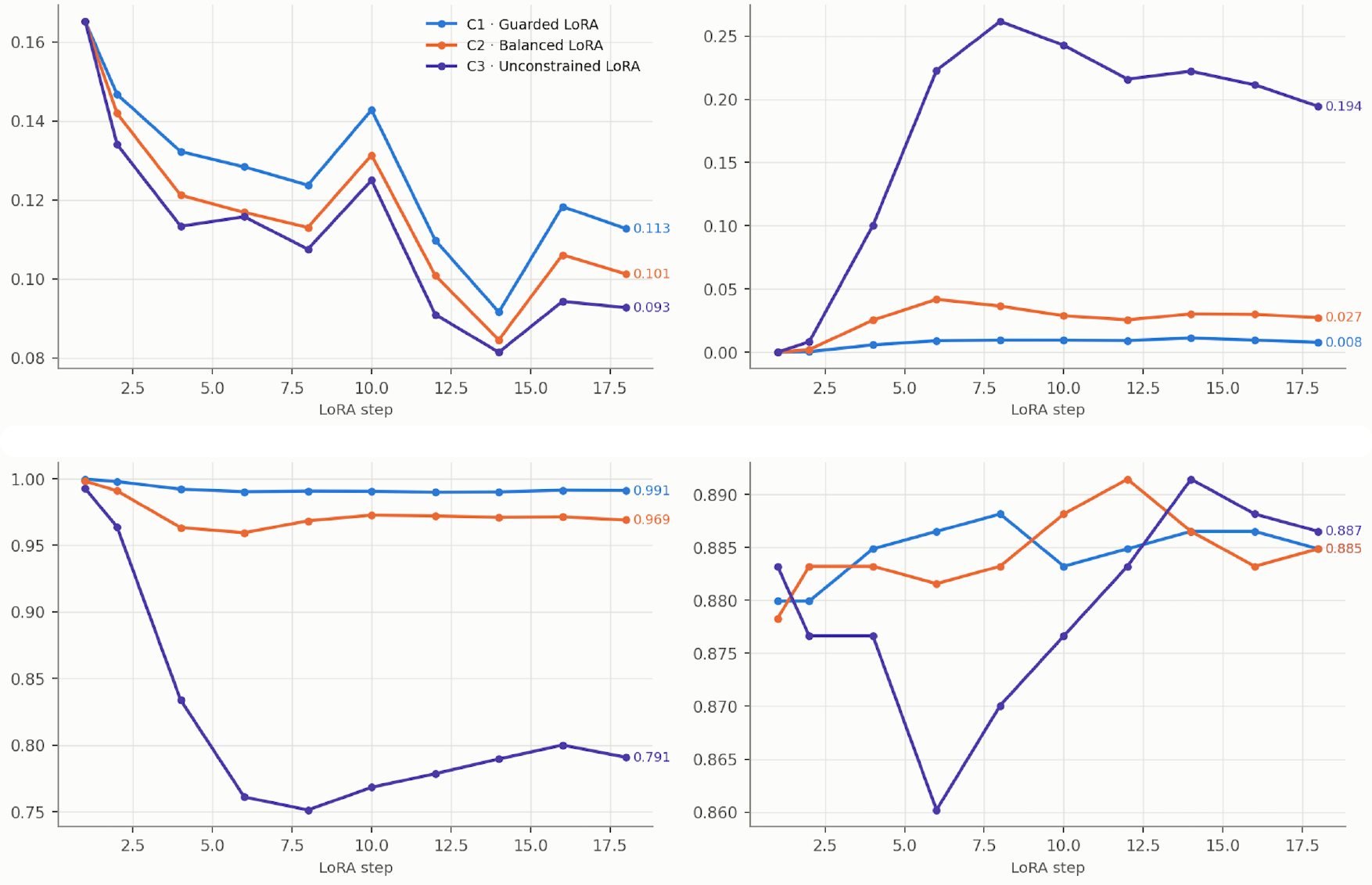}
\caption{\small Training dynamics of the controlled sweep. \textbf{Top left:} BCE alignment loss of olfaction and text. \textbf{Top right:} anchor distillation loss. \textbf{Bottom left:} anchor drift. \textbf{Bottom right:} R@5 held-out loss over olfaction-text. The distillation loss cleanly separates the guardrail regimes. C1 stays pinned, C3 wanders, and the held-out R@5 for C3 briefly dips while its geometry reorganizes, then recovers.}
\label{fig:embgeom-loss}
\end{figure}

Results for the ablation study can be observed in Table \ref{tab:ablation}.
The 18-step budget is another form of control, held equal across C1--C3 so the $\lambda$-guardrail and LoRA adaptor capacity are the only variables. 
It is fair to ask whether the finetunes were stopped early, especially since the C1--C3 t-SNE panels look so similar. 
To evaluate this, we re-run all three finetunes to 48 steps (same seed, same descriptor minibatches), tracking  anchor drift and olfaction-language alignment quality.

\begin{figure}[h]
\centering
\includegraphics[width=0.98\linewidth]{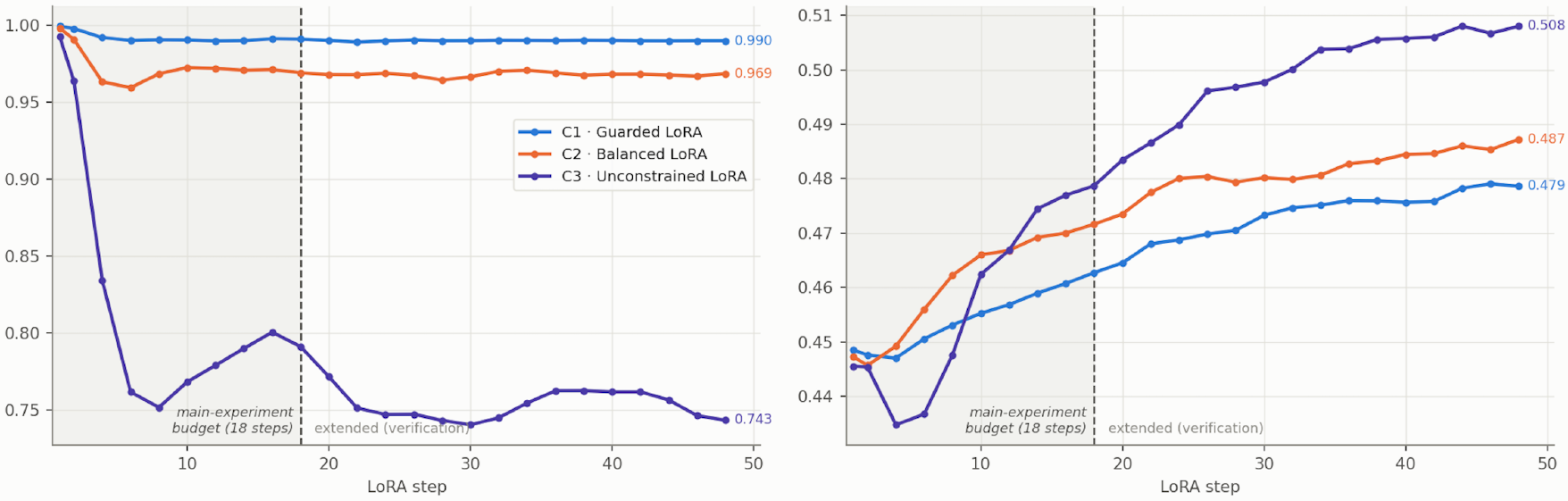}
\caption{\small Convergence check: an extended 48-step re-run of the three controlled finetunes. \textbf{Left:}
anchor drift (what the t-SNE visualizes) is essentially flat after step $\sim$6, so more steps would not
move the panels. \textbf{Right:} olfaction$-$language mAP saturates, with only a small residual gain
past the 18-step budget (dashed line). The near-identical C1-C3 t-SNE panels reflect converged geometry, not
early stopping.}
\label{fig:embgeom-conv}
\end{figure}

Over the 30 extra steps, we found that the geometry had converged, but the alignment had not.
Drift moved by $-0.001$ for C1 ($0.991\to0.990$) and $-0.001$ for C2 ($0.969\to0.969$), and effectively remained flat after step $\sim$6, so their
t-SNE panels would be pixel-identical at 48 steps. 
C3, the unguarded case, creeped from $0.791\to0.743$, $-0.048$ over 30 steps and stayed inside the same attractor basin. 
This contrast underscores the whole point: 30 extra steps of the gentle schedule moved C3's drift by $\sim\!0.05$; a single step of C4's force moved drift by $\sim\!0.99$ ($1.000\to0.005$). 
At the C1-C3 learning rate, the geometry is pinned near frozen no matter how long the model is trained; only more force via hotter LR, looser clip, or higher $\lambda$ relocates it. 
Meanwhile mAP kept inching up (C1 $+0.016$, C2 $+0.015$, C3 $+0.029$ to $0.508$), because the adapters make small in-place refinements at roughly constant drift. 
So the near-identical C1-C3 panels reflect converged geometry, \textit{not} early stopping.
Ttraining longer would improve alignment slightly without moving the t-SNE, which is precisely why the geometry plot is the right lens for the guardrail's effect and the wrong lens for the extra alignment a granted by a longer training schedule.

Final results of the ablation study show that guarded finetuning \emph{strengthens} bidirectional olfaction-language monotonically.
Across the controlled sweep, every finetune beat off-the-shelf on the ranking metrics (mAP $0.449\to0.479$, $+6.7\%$ relative; R@10 $0.911\to0.934$).
The observed gain grew as the guardrail relaxed, with the alignment loss moving the descriptor anchors to fit the fixed olfaction cloud better.
More freedom means a tighter fit. 
With this, we have evidence to suggest that off-the-shelf models and data are a floor, and finetuning helps when cleanly architected.

We find that the $\lambda$-guardrail is a seemingly reliable tuning dial on geometry drift. 
With the configurations involving the guardrail, the olfaction$-$vision grounding appeared self-preserving with high inertia to movement.
Even at C3's substantial drift, text-image properties remain relatively unchanged.
The alignment objective pulls the anchors toward the olfaction cloud, which the encoder placed near the original image-aligned geometry.
Consequently, the objective that strengthens language also preserves vision. 
There is no gentle ``vision breaks first'' regime in C1--C3.

Conversely, we find that this preservation has a limit.
By removing the $\lambda$-guardrail in C4 and supplying enough force, the objective can no longer hold the encoder together, and it collapses in a single step. 
Everything routed through the text encoder of the VLM collapses at once - olfaction-language falls to near-random (R@1 $0.627\to0.025$, mAP $\to0.062$), descriptor-image grounding breaks (category top-1 $0.542\to0.167$, the random baseline), and the general text-image ceiling collapses ($0.938\to0.062$).
Essentially, the text encoder forgot how to map any text to their rightful images, and anchor drift hits $0.005$ as a result. 
The one number that does \emph{not} move is the direct (z vs image) control at $0.583$, because it never passes through the finetuned text encoder, suggesting that the failure may be localized entirely to the broken text encoder. 
Finally, we find that grounding does not degrade gracefully - it survives until the encoder catastrophically collapses, then goes to random with everything else.
        
Robots utilizing olfactory hardware will ground some type of gas sensor (e.g. electrochemical, metal-oxide, optical, acoustic), not a molecule.
The on-device heatmap follows the identical text-image geometry probed here, and accuracy of the heatmap can be affected by the language-side retrieval tuning above.
A guarded finetune can sharpen the language-side retrieval the sensor encoder trains against without disturbing the sensor$-$image grounding that powers the heatmap, whereas an over-finetune destroys the ceiling and would send sensor$-$image retrieval to random. 
The $\lambda$-guardrail is what keeps LoRA adaptation from catastrophically collapsing as seen in the C4 configuration.

In summary of the ablation study above, the $\lambda$-guarded C1 configuration captured most of the language-side gain (R@5 $+0.7$\,pt, mAP $+1.4$\,pt, R@10 $+1.0$\,pt) at negligible drift ($0.991$) and zero grounding cost. 
C2 and C3 were good regressive research probes into the effects of variations of LoRA and the $\lambda$-guardrail.
C4 is the boundary condition showing exactly what an unguarded, over-aggressive finetune does by collapsing to randomness.

\section{Federated Learning \& Privacy}\label{sec:federated_learning}
The results from Section \ref{sec:embgeom} demonstrate the limits of what current open source datasets can provide toward enabling models to learn olfaction-vision relationships.
To provide better olfactory intelligence for robotics, more multimodal data must be gathered.
Scentience has developed the infrastructure to acquire this data and is using it today to aggregate olfaction-vision data \textit{en masse} through the deployment of its own olfactory instruments in several labs, companies, and universities around the world.
One of the most critical components of this infrastructure is the \textit{Federated Learning and Observability System (FLOS).}
Scentience uses federated learning to enable each olfactory instrument to learn from the others within the fleet, and to continually appreciate the accuracy of its models as instruments gather out-of-distribution data. 

At a high level, the FLOS has three parts. On each device, a local model builds up its own understanding of the olfactory patterns it observes during normal operation. In the cloud, a global model combines the distilled latents from many devices - never their raw data. 
The global model then retrains and broadcasts an updated checkpoint back out to each node in the fleet. 
Underneath both sits a privacy architecture that keeps identifiable raw data strictly separated from the de-identified copy the learning system is allowed to touch.

The learning loop itself is analogous to a state machine: aggregate what the devices have learned, re-train the global model, and ground it against trusted benchmarks. Those steps repeat perpetually, tightening the alignment between the grounding benchmarks and the global model over time.

\begin{figure}[t]
\centering
\includegraphics[width=0.82\linewidth]{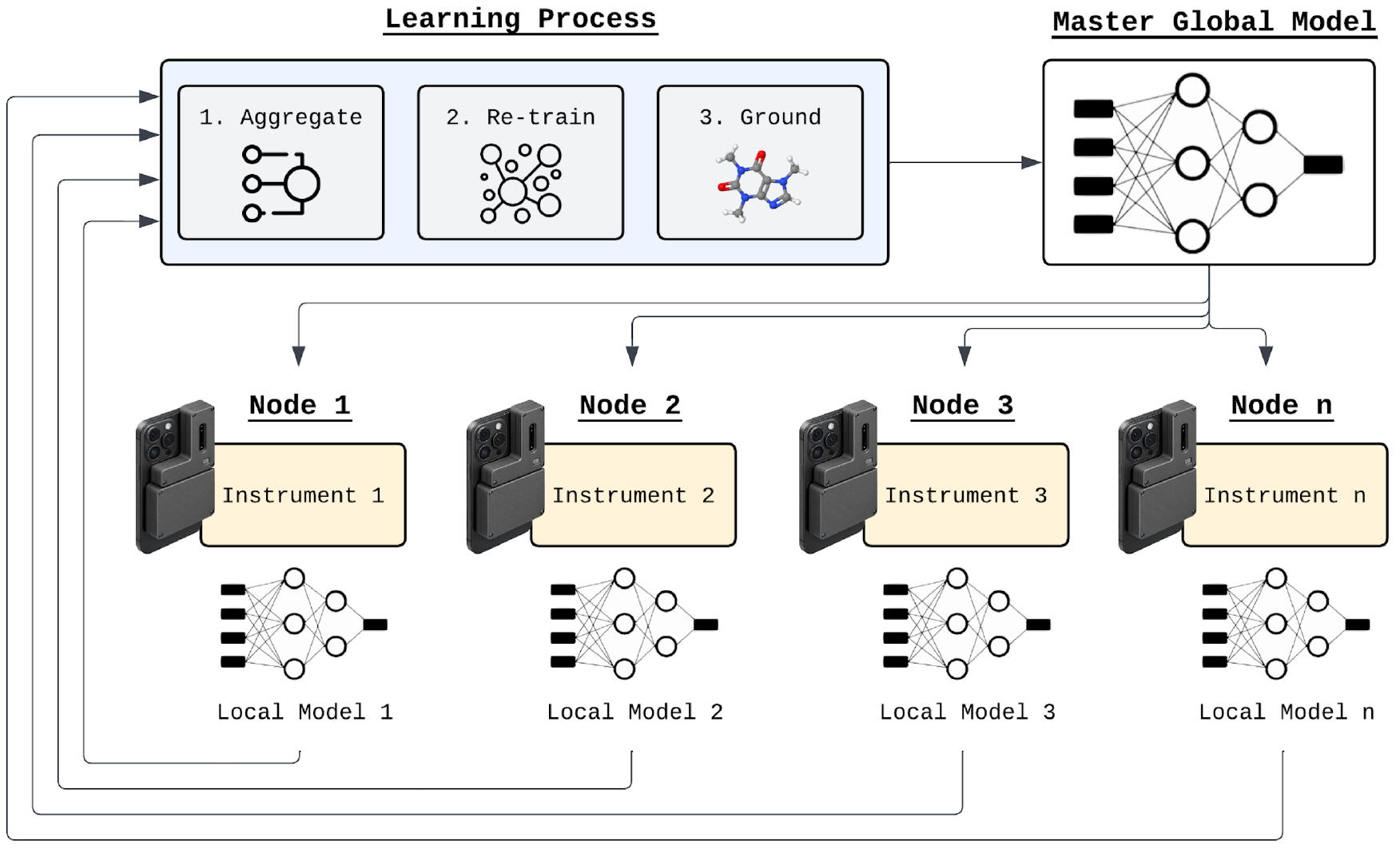}
\caption{\small Scentience's models and sensors are controlled and calibrated through continuous learning via federated weight updates.}
\label{fig:sigma_fl}
\end{figure}

The manner in which each device manages memory underscores the notion of privacy.
Figure \ref{fig:sigma_fl} illustrates how on-device models ground themselves with the cloud.
Every on-device model draws on three kinds of memory at once: a local memory of what it has recently observed itself, a global memory pulled down from the cloud, and an external memory that lets it see how similar patterns look on its peer devices. 
Each plays a distinct role. 
Local memory captures what an instrument sees most reliably in its own environment. 
Global memory pools cross-device evidence so that one device’s local nuances never hardens into doctrine. 
External memory lets a device hold its own beliefs up against the broader system before trusting them in the manner of Prototypical Networks from \cite{snell2017prototypical}.

Through this tricycle design, all three aspects contribute to a successful and grounded continually-appreciating intelligence. 
Federated learning in olfaction should not merely average gradients - it should preserve and compare structured insights from the fleet that allow for alignment on deeply personal experiences that maintain the warranted privacy.

\subsection{Learning Without Labels}

One of the hardest constraints is the absence of a labeled training set. 
The FLOS gets around this by having each device learn primarily from its own most confident observations. 
In the spirit of prototypical networks from \cite{snell2017prototypical}, when the model is confident about what it is seeing, that observation becomes a reliable internal teaching example that establishes a prototype. 
Analogously, when the model is unsure, that observation is treated as a low-value outlier and its training emphasis is reduced in the distribution. 
Over time, each device distills its experience down toward the cleanest, most trustworthy examples of each relationship it knows, building its own training signal out of its own confidence without dependence on a human to label everything.

\subsection{Guarding Against Bad Prototypes}

A device with a failing sensor, an unusual local environment, or a skewed drift pattern can send up a misleading summary and establish bad prototypes. 
To guard against this, when a device contributes, the cloud sends back a small ranked set of the most similar summaries from its peers. 
The device can compare its own view against each peer set before fully committing to an update - a “consensus check” of sorts. 
A prototype that disagrees sharply with everything its peers are seeing is treated as suspect. 
Because only a small number of peer summaries come back, this keeps compute cheap even as the deployed fleet grows. 
Guarding against bad prototypes is difficult enough with one modality and this is only exacerbated among multimodal data.
Our model infrastructure provides a series of checks and balances to protect against and catch bad prototypes.

\subsection{Keeping Identifiable Data Walled Off}

Privacy is in the FLOS DNA. 
The privacy-oriented architecture splits the data path in two. 
One store holds the original, encrypted, identifiable assessment from the device and is never touched for training - it is read only when the user (or someone they authorize) explicitly asks for their own data. 
A second store holds a de-identified working copy with identifying information stripped out, and that is the only thing on which the learning system is allowed to train.
The permissions between the two are deliberately asymmetric and enforced in the architecture itself versus being left to policy: the learning pipeline can read the de-identified store but not the identifiable one, and the user-facing app can write to the identifiable store but not its de-identified counterpart.
\section{Limitations \& biases}\label{sec:limitations}
While we consider \textit{COLIP-2} to be the state of the art for olfaction-vision-language modeling, there are a few honest limitations of which users should be aware. 
We detail the primary limits and failure modes captured in our testing battery in the following sections.
For each limit and failure mode noted, Scentience is actively working with communities across robotics, perfumery, food chemistry, animal science, biochemistry, government policy, and neuroscience to rectify for the next model.
\begin{itemize}[leftmargin=1.2em,itemsep=2.5pt,topsep=2pt]
  \item \textbf{Grounding is family-level, and partly emergent.} Olfaction-image
        alignment rides on the VLM's text-image space and is reliable at the \emph{odor-family} level, not the single-molecule level. It is sharper from a whole-substance sensor signature than from a single molecule. 
  \item \textbf{Novel substances are not identified exactly.} The encoders learn discriminative
        signatures for \emph{trained} classes; genuinely unseen aromatic patterns transfer only weakly, at the odor-family level. True zero-shot identification is clearly a natural next step for future work.
  \item \textbf{Small, biased data.} A few thousand molecules and a modest sensor corpus; generalization to all chemistry domains for all olfactory sensor types is not possible with \textit{COLIP-2}. We integrated the most common chemical sensors used by researchers and roboticists today over the small open-sourced olfactory datasests available.
  \item \textbf{Perceptual subjectivity.} Odor perception is panel- and concentration-dependent and reflects the (often Western, expert or untrained) human panels behind the source lexicons; hedonic axes such as pleasantness and intensity are deliberately not modeled. More accurate taxonomies are being developed by companies like \textit{Osmo}, \textit{Monell}, perfumeries, and others, and we intend to build off these as they mature.
  \item \textbf{Outputs are rankings, not calibrated probabilities.} Descriptor scores, retrieval similarities, and heatmap intensities are useful for ordering and localization, not as exact representations of single molecules or precise concentrations.
\end{itemize}

\subsection*{Known failure modes}
Known measured failure modes that exist at the aroma-family level captured during our evaluation battery are presented in Table \ref{tab:limits}.

\begin{table}[h]
\caption{Known Failure Modes}
\label{tab:limits}
\vspace{2mm}
\centering\small\sans
\begin{tabular}{L{5.2cm} L{11.0cm}}
\hdrow \thd{Failure mode} & \thd{Behavior} \\
Single molecule $\rightarrow$ image is weak
  & A single character-impact compound grounds far less reliably than a full sensor signature (exact top-1 $\approx$ 0.16 vs 0.72); in other words, one compound underspecifies a food's full aroma. \\
\rowcolor{scrowalt} Odor-adjacent visual confusions
  & An aroma retrieves odor-similar foods first (e.g.\ a ``lemon'' query surfaces other citrus/spice items before lemon); ranking is family-level, not exact. \\
Novel substances not identified
  & Substances never observed in training are not identifiable at the single-odor level and maintain only weak family-level transfer. \\
\rowcolor{scrowalt} Heat spreads on smell-similar scenes
  & When several odor-similar objects share a frame, the heatmap distributes across them; this should be interpreted as a probability field, not a precise detector. \\
Out-of-domain gases
  & Inorganic gases (e.g.\ NH$_3$, NO, CO) have no representation in a food-odor model and are surfaced as out-of-distribution rather than forced into a confident match. \\
\end{tabular}
\end{table}
\section{Ethics, safety \& privacy}
\textit{COLIP-2} is a research prototype. 
It must not be used as a safety device (leak or toxicity detection) or for medical or diagnostic purposes without dedicated validation; an odor match says
nothing about precise concentration, toxicity, or safety. 
The source lexicons on aroma data carry the cultural and methodological biases of the specific human panels behind them.

The system can ingest camera images and gas-sensor readings, so data handling is a first-class concern. 
The on-device paths keep data local: molecule, sensor embedding, and image embedding all run on-device with no network egress. 
Hosted grounding transmits the raw image to the server for encoding; because such images may contain faces, people, locations, or other personal information, users of the model API are responsible for consent and for preferring the on-device path when it suffices. 
Where the architecture allows, model improvement uses privacy-preserving federated learning that enable model updatse to occur via on-device encoders rather than raw data. 

\section{Conclusion}
Our goal with \textit{COLIP-2 } is to define a solid architecture trained on open datasets that shows the limits on how robust olfaction-vision-language relationships can be learned.
While useful in its own right, we encounter obvious barriers warranting the accumulation of more data.
These restrictions are not inherently due the architecture's ceiling, but rather that no architecture can synthesize olfaction–vision relationships that have never been observed.
Furthermore, heterogeneous sensor outputs and the lack of ImageNet-scale olfactory data limits the ability to conduct sensitive and precise odor localization.
Models and data are saturated on the vision axis, but the next order‑of‑magnitude gain will come from acquiring paired olfaction–vision data \textit{en masse}, not from a better loss function or a bigger VLM.

Scentience releases \textit{COLIP-2} as a public baseline while we build the sensors, silicon, infrastructure, and data pipelines that will construct a more robust \textit{COLIP-3} in collaboration with scientists across several disciplines in academia and industry.
\section{License}\label{sec:license}

This document is publicly released. 
Model weights, training code, production serving artifacts, taxonomy-generation methods and selected data-processing pipelines are not publicly distributed. 
Hosted and research access may be made available by Scentience under separate terms.
Scentience owns its original model components, trained weights, taxonomy, sensor-alignment methods, software, documentation and serving artifacts, subject to applicable rights and restrictions associated with third-party components and source datasets.

\section{Acknowledgments}
\label{sec:ack}
We would like to thank the following individuals for their feedback on \textit{COLIP-1} and guidance on how to construct multi-modal models that are aligned to human perception (ordered alphabetically by last name): Ovidiu Daescu (UT Dallas), Nik Dennler (ETH Zürich), Han Fan (TU Munich), Matthew Kilcoyne (independent), Michelle Niedziela (Nerdoscientist, LLC),  Rohith Peddi (UT Dallas), Robert Pellegrino (Texas A\&M University), and Tian Yu (AMAI Consulting).


\vspace{10pt}
\hrule height 0.4pt
\vspace{4pt}
\

\bibliography{main}
\bibliographystyle{plain}

\end{document}